\documentclass{article}
\usepackage{graphicx} % Required for inserting images
\usepackage{authblk}
\usepackage{algorithm}
\usepackage{algorithmic}
\usepackage{amsmath}
\usepackage{amsfonts}
\usepackage{booktabs}
\usepackage{multirow}
\usepackage{subcaption}
\usepackage{tabularx}
\usepackage{url}
\usepackage{hyperref}
\usepackage[style=vancouver, natbib=true, backend=bibtex]{biblatex}
\usepackage[section]{placeins}
\usepackage[top=2cm, bottom=2cm, left=2cm, right=2cm]{geometry}
\bibliography{bibliography.bib}
\title{Neuromorphic Astronomy: An End-to-End SNN Pipeline for RFI Detection on Hardware}
\author[1,2]{Nicholas J. Pritchard}
\author[1]{Andreas Wicenec}
\author[1]{Richard Dodson}
\author[2]{Mohammed Bennamoun}
\author[2]{Dylan R. Muir}

\affil[1]{International Centre for Radio Astronomy Research, University of Western Australia, Australia}
\affil[2]{School of Physics, Mathematics and Computing, University of Western Australia, Australia}
\date{November 2025}

\begin{document}
\hypersetup{hidelinks}
\maketitle
\begin{abstract}
Imminent radio telescope observatories provide massive data rates making deep learning based processing appealing while simultaneously demanding real-time performance at low-energy; prohibiting the use of many artificial neural network based approaches.
We begin tackling the scientifically existential challenge of Radio Frequency Interference (RFI) detection by deploying deep Spiking Neural Networks (SNNs) on resource-constrained neuromorphic hardware.
Our approach partitions large, pre-trained networks onto SynSense Xylo hardware using maximal splitting, a novel greedy algorithm.
We validate this pipeline with on-chip power measurements, achieving instrument-scaled inference at 100mW.
While our full-scale SNN achieves state-of-the-art accuracy among SNN baselines, our experiments reveal a more important insight that a smaller un-partitioned model significantly outperforms larger, split models.
This finding highlights that hardware co-design is paramount for optimal performance.
Our work thus provides a practical deployment blueprint, a key insight into the challenges of model scaling, and reinforces radio astronomy as a demanding yet ideal domain for advancing applied neuromorphic computing.
\end{abstract}
\section{Introduction}
Increasing demand for neural-network-based Artificial Intelligence (AI) across diverse domains is driving significant growth in data-centre energy usage, underscoring a need for effective, widely deployable, and efficient computing \cite{iea_energy_2025, muir_road_2025}.
Current computational approaches meet raw processing requirements, but doing so with lower resource consumption through selective borrowing of concepts from neuroscience remains a promising and active area of research \cite{muir_road_2025}.
Radio Frequency Interference (RFI) detection in radio astronomy exemplifies this challenge.
RFI consists of anthropogenic radio emission that overwhelms or corrupts the faint cosmic signals telescopes are designed to capture, making accurate and efficient detection critical for the integrity of radio astronomy \cite{thompson_interferometry_2017, van_nieuwpoort_towards_2016}.
RFI is becoming increasingly prevalent, colliding with the increasing sensitivity of future instruments, making RFI detection an existential challenge for the field \cite{noauthor_report_2022}.
Traditional RFI detection algorithms and recent machine-learning-based approaches often treat the problem as an image-segmentation problem suitable for post-observation batch processing \cite{offringa_aoflagger_2010, mesarcik_learning_2022}.
Current operational algorithms require expert calibration, which will not scale to meet the data demands of incoming instruments.
Additionally, ANN approaches remain operationally expensive to deploy practically \cite{dutoit_comparison_2024}, as many model instances are required to run in parallel to service an entire instrument, which produces a spectrogram per pair of antennae (baseline), which for contemporary large interferometric telescopes numbers in the tens of thousands \cite{vermij_challenges_2015}.

Spiking Neural Networks (SNNs) offer a compelling alternative to conventional Artificial Neural Networks (ANNs) by drawing their communication mechanism from biology; transmitting discrete, event-based `spikes' rather than continuous weighted sums, with each neuron exhibiting time-varying internal dynamics, lending themselves to temporal information processing \cite{trappenberg_fundamentals_2010}.
Neuromorphic computers are specialised hardware platforms designed to execute SNN models directly.
When the task and input data characteristics align with the time-varying dynamics of SNNs, these systems can yield immense efficiency gains \cite{schuman_opportunities_2022} but only through careful consideration of a target task and spiking dynamics \cite{ottati_spike_2023, dampfhoffer_are_2023}.
Recent advancements, including reliable backpropagation-like training rules for SNNs \cite{neftci_surrogate_2019}, and sophisticated open-source tooling \cite{eshraghian_training_2023, muir_rockpool_2019, pedersen_neuromorphic_2024}, have spurred a renaissance in the field.
These advancements have shifted the focus from biological modelling towards addressing practical, high-impact applications such as autonomous driving \cite{zhu_autonomous_2024}, language modelling \cite{zhu_spikegpt_2023}, audio tagging \cite{bos_sub-mw_2022}, gesture recognition \cite{massa_efficient_2020}, robotic control \cite{glatz_adaptive_2019} and olfactory sensing \cite{dennler_rapid_2022}.

Despite these promising developments and the potential of SNNs, the development and deployment of complex, state-of-the-art SNN models onto currently available, often resource-constrained neuromorphic hardware, characterised by finite neuron counts, strict connectivity limitations, and memory constraints, remains a significant challenge \cite{schuman_survey_2017, schuman_opportunities_2022}.

SNNs have previously demonstrated compelling RFI detection performance \cite{pritchard_spiking_2024}, yet a gap exists in translating high-performing models to real-world neuromorphic hardware.
To begin addressing this challenge and exploring the viability of SNNs for this scientific application, we make the following contributions:
\begin{enumerate}
    \item We develop a Backpropagation-Through-Time (BPTT) trained SNN that achieves high accuracy on a synthetic RFI detection benchmark, surpassing previous algorithmic and SNN baselines and competitive with ANN baselines. 
    \item We propose `maximal splitting', a novel greedy algorithm designed to shard large pre-trained SNNs onto hardware with specific neuron, synapse, and fan-in constraints.
    \item We demonstrate a complete end-to-end pipeline from SNN model training in snnTorch, translating to Rockpool using the Neural Intermediate Representation framework (NIR) \cite{pedersen_neuromorphic_2024} then deploying split models on the SynSense Xylo 2 neuromorphic processor.
\end{enumerate}
We critically evaluate the maximal splitting method, comparing performance against naive and random splitting baselines, and against small SNN models trained from scratch specifically for direct hardware deployment, and expose a crucial flexibility-versus-efficiency trade-off.
Through these efforts, we further validate RFI detection as a demanding yet relevant benchmark for advancing neuromorphic computing research and hardware capabilities, highlighting the need for continued research into hardware-aware training methodologies and deployment pipelines to unlock scalable and high-performing SNN-powered solutions.
\section{Related Work}\label{sec:related}
\paragraph{Spiking Neural Networks}
Spiking Neural Networks (SNNs) represent a class of neural network drawing inspiration from computational principles of biological nervous systems \cite{trappenberg_fundamentals_2010, maass_networks_1997}.
The core computational unit in SNNs, the spiking neuron, can be modelled in various ways, differing in computational efficiency and biological realism \cite{gerstner_spiking_2002}.
An increase in complexity makes spiking neurons more expressive \cite{maass_networks_1997}, but also more difficult to train \cite{neftci_surrogate_2019}.
The field has spent decades exploring biology-inspired learning rules \cite{taherkhani_review_2020} however, Backpropagation Through Time (BPTT) with surrogate gradients \cite{neftci_surrogate_2019} has created somewhat of a watershed moment, as complex SNNs can be trained with increasing reliability and a new generation of neuromorphic hardware promise to unlock previously impossible levels of efficiency in neuron-based processing \cite{muir_road_2025}.
\paragraph{Neuromorphic Hardware Aware Training}
SNNs and neuromorphic computing platforms have developed closely together with their origins rooted in the use of sub-threshold analog circuits to emulate neuron dynamics \cite{mead_neuromorphic_1990}.
Since then, `neuromorphic' hardware has grown to include a wide variety of approaches spanning mimetic biologically-inspired approaches to top-down fully digital architectures \cite{schuman_survey_2017, frenkel_bottom-up_2023}.
Unlike readily accelerated ANNs, SNNs are often paired with specialised low-precision analog or digital circuits.
Contemporary neuromorphic systems impose strict deployment constraints, including weight precision, memory and communication budgets, neuron counts, connectivity limits, and analog hardware mismatches, making practical SNN-based application deployment difficult.

Earlier attempts to deploy SNNs circumvented these issues via ANN-to-SNN conversion strategies \cite{esser_backpropagation_2015, rueckauer_conversion_2017}.
Recent research has moved towards hardware-aware training, where the target system's constraints are incorporated.
These include quantisation-aware training for digital chips like Intel's Loihi \cite{davies_loihi_2018}, as well as analog-in-the-loop methods running a forward pass in hardware but refining model parameters offline \cite{cramer_surrogate_2022}.
Finally, emulating larger-than-hardware models sequentially remains a flexible option \cite{arnold_scalable_2025}.
Despite progress, most approaches treat SNNs as monolithic models to be mapped entirely to hardware.
As neuromorphic hardware and training methodologies mature, new questions emerge around how to co-design scalable, efficient, accurate and deployable models bridging abstract learning paradigms with the realities of hardware-level execution.
\paragraph{RFI Detection}
RFI detection and mitigation are long-standing critical processes in radio astronomy.
Traditional methods rely on fast cumulative-sum-based algorithms such as AOFlagger \cite{offringa_aoflagger_2010}, requiring expert tuning to `flag' contaminated `visibilities' in spectrograms.
While operationally effective, manual tuning for specific instruments and observing conditions makes these methods difficult to use for incoming massive radio telescopes such as the Square Kilometre Array (SKA) \cite{vermij_challenges_2015}.

The existence of large astronomical datasets and the need for data-driven approaches to RFI detection have encouraged the development of AI-based RFI detection methods.
Convolutional Neural Networks (CNNs) and U-Net architectures, in particular, have been extensively explored for this task, treating RFI detection as a two-dimensional semantic segmentation task, just as humans and traditional algorithms do \cite{akeret_radio_2017, mosiane_radio_2017, vos_generative_2019, vafaeisadr_deep_2020, yang_deep_2020}.
Employing anomaly detection schemes has also been investigated, utilising auto-encoders \cite{mesarcik_learning_2022, vanzyl_remove_2024}.
Most recently, authors have explored segment anything networks \cite{deal_segmenting_2024} and vision transformers \cite{ouyang_hierarchical_2024}.
These approaches have shown promise, albeit with varying trade-offs in performance, but remain operationally expensive to deploy \cite{dutoit_comparison_2024}.
The need for real-time RFI detection has spurred the field to explore RFI detection earlier in the signal chain with shorter time-windowed cumulative sum-based approaches \cite{mehrabi_real-time_2024}.

The spectro-temporal nature of radio astronomy visibilities makes it an appealing domain for SNNs \cite{scott_evolving_2015, kasabov_evolving_2016}.
Initial explorations into SNN-based RFI detection adapted an anomaly detection scheme through ANN-to-SNN conversion \cite{pritchard_rfi_2024}.
Subsequent work focused on directing training SNNs through BPTT by reformulating RFI detection as a time-series segmentation task \cite{pritchard_spiking_2024, pritchard_supervised_2024}.
Other work explores using liquid state machines for this task with limited success \cite{pritchard_advancing_2025}.
While still an emerging area, SNN-based methods hold the potential to combine neural networks' pattern recognition capabilities with low-power, event-driven processing ideal for real-time applications on neuromorphic hardware.
\section{Methodology}\label{sec:xylo:methods}
\subsection{RFI Detection with SNNs}
We briefly present the preliminaries for RFI detection as a time-series segmentation problem, aligning with existing work in the literature \cite{mesarcik_learning_2022, pritchard_rfi_2024, pritchard_spiking_2024}.
Radio interferometer array observatories process the raw voltages from antennae into complex-valued `visibility' data $V(\upsilon, T, b) \in \mathbb{C}$ varying in frequency, time, and baseline (pair of antennae).
RFI detection involves producing a boolean mask of `flags' $G(\upsilon, T, b) \in \{0, 1\}$ varying in the same dimensions.
The time-series segmentation formulation to RFI detection is given as
\begin{equation}\label{eq:xylo:newformulation}
    \mathcal{L}_{\text{sup}} = 
    min_{\theta_n}(
        \Sigma_t^T
        \mathcal{H}(
            m_{\theta_n}(
                E(V(\upsilon, t, b), e)
            ), 
            F(G(\upsilon, t, b), e)
        )
    )
\end{equation}
where $\theta_n$ are the parameters of some classifier $m$, $\mathcal{H}$ is a similarity measure, $E$ is an input encoding function and $F$ is an output encoding function.
Both functions introduce a new integer exposure parameter, $e$, controlling the spike train length for each time step in the original spectrogram.
The encoding function results in a time-series of dimensions $(\upsilon, T \cdot e, b)$ effectively presenting each time-step in the original spectrogram for an extended period of time $e$.

We use latency encoding and decoding in this work based on previous works that found this method to be most effective at this task which we present here for completeness \cite{pritchard_supervised_2024}.
For a given exposure $E$, input intensities are mapped inversely and linearly from 0 to $E$.
Output decoding interprets spikes before the last exposure step as RFI and all other inputs (including silence) is background.
The following equation maps the spike times in the supervised masks:
\begin{equation}\label{eq:latency:mask}
    t = \begin{cases}
        0 & G(\upsilon, t, b) = 1 \\
        E & otherwise
    \end{cases}
\end{equation}
The comparison function, $\mathcal{H}$, is the mean square error of spike timings given by the following function:
\begin{equation}\label{eq:xylo:latency:comparison}
    \mathcal{H}_{\text{latency}} = \Sigma_e^E\Sigma_\upsilon^\Upsilon(y_{\upsilon,e} - f_{\upsilon, e})^2.
\end{equation}

\paragraph{Fan-in Aware Regularisation}
We encourage accurate model partitioning by regularising the network to respect the fan-in requirements of the target hardware platform.
The penalty term calculates the average excess fan-in, $p_{f_{in}}$, for each neuron in all layers as per,
\begin{equation}\label{eq:xylo:penalty}
    p_{f_{in}} = \frac{1}{f_{in}} \cdot \frac{1}{N} \sum_{i=1}^{N} \max\left( \sum_{j=1}^{M} \mathbb{I}(|w_{ij}| > \varepsilon) - f_{in},\ 0 \right)
\end{equation}
where $\epsilon$ is a floating-point limit ($1e-8$), $i$ iterates over each layer in the network, $j$ iterates over each neuron in each layer, $\mathbb{I}$ is the indicator function and $f_{in}$ is the maximum fan-in for the target hardware platform (in this case 63 for the SynSense Xylo 2).
\paragraph{Loss Function}
Finally, the loss function becomes
\begin{equation}
    \mathcal{L} = \mathcal{L}_{\text{sup}} + \lambda \cdot p_{f_{in}}
\end{equation}
where $\mathcal{L}_{\text{sup}}$ is the supervised loss function from Equation \ref{eq:xylo:newformulation}, $\lambda$ is a hyper-parameter controlling the input of the fan-in penalty and $p_{f_{in}}$ is the fan-in penalty defined in Equation \ref{eq:xylo:penalty}.
The result is a robust formulation to RFI detection as a time-series segmentation problem with latency encoding and hardware fan-in consideration for SNN training.
\subsubsection{Spiking Neuron Model}
This work employs a second-order Leaky Integrate and Fire (LiF) neuron model parameterised by $\tau_{\text{syn}}$ (equivalently $\alpha = e^{(-\Delta t / \tau_{syn})}$ in snnTorch), a decay rate for the synaptic decay, and $\tau_{\text{mem}}$ (equivalently $\beta = e^{(-\Delta t / \tau_{mem})}$ in snnTorch), a decay rate for the membrane decay.
Including both membrane and synaptic decays enables this neuron model to more easily sustain inputs for longer periods, boosting capability in our latency-driven approach \cite{eshraghian_training_2023}.

There exists subtle differences in how snnTorch, Rockpool and NIR model the second-order LiF neuron but all are parameterised by the same variables $\tau_{\text{syn}}$, and $\tau_{\text{mem}}$, in addition to $\Delta t$ which controls the libraries' simulation granularity.
SnnTorch models the neuron with the following equation:
\begin{equation}
\begin{aligned}
    I_{\text{syn}}[t+1] = e^{\frac{-\Delta t}{\tau_{\text{syn}}}} I_{\text{syn}}[t] + WX[t+1] \\
    V_{\text{mem}}[t+1] = e^{\frac{-\Delta t}{\tau_{\text{mem}}}} V_{\text{mem}}[t] + I_{\text{syn}}[t+1] - R[t]
\end{aligned}
\end{equation}
Where $I_{\text{syn}}$ and $V_{\text{mem}}$ is the synaptic current and membrane potential respectively, $X$ is the input current, $W$ is an incoming weight and $R[t]$ is a reset mechanism, resetting the neuron's membrane potential by subtraction upon spiking.

Rockpool models the neurons with the following equations:
\begin{equation}
\begin{aligned}
    I_{syn}^{t+1} = I_{syn}^{t} + S_{in}(t) + S_{rec} \cdot
    W_{rec} \\
    I_{syn}^{t+1} = I_{syn}^{t} \cdot e^{(-\Delta t / \tau_{syn})} \\
    V_{mem}^{t+1} = V_{mem}^{t} \cdot e^{(-\Delta t / \tau_{mem})} \\
    V_{mem}^{t+1} = V_{mem}^{t} + I_{syn}^{t} + b + \sigma \zeta(t)
\end{aligned}
\end{equation}
as per Rockpool's implementation \cite{muir_rockpool_2019} where $V_{mem}$ is the membrane voltage potential, 
$I_{syn}$ is the synaptic current,
$S_{in}$ are the incoming spikes,
$S_{rec}$ are recurrent spikes,
$W_{rec}$ is the current weight,
$\tau_{syn}$ is the synaptic time-constant,
$\tau_{mem}$ is the membrane time-constant,
$b$ is a bias current,
$\zeta(t)$ is a Wigner random noise function, and the neuron membrane is reset by subtraction.
While structurally similar to snnTorch's implementation, the change to multiply the updated state variables by a decay rate, and including bias and noise perturbation means there is not a guaranteed direct correspondence between these libraries.
This order of operations matches that found in the Xylo hardware.
Finally, NIR represents a second-order LiF neuron with a Current-based activation LIF model (CubaLiF), which is itself composed of a first-order LiF neuron followed by a linear layer and a final leaky integrator discussed most clearly in the original text \cite{pedersen_neuromorphic_2024}.

Notably, the decay parameters written out by snnTorch to NIR get read in to Rockpool without storing the $\Delta t$ value which is instead dynamically computed, the result are $\Delta t$ values which vary within the read network by default, something we manually correct when reading the NIR model into Rockpool. 

\subsection{Framework Conversion and Model Architecture}
\begin{figure}[!htb]
    \centering
    \includegraphics[width=1.0\linewidth]{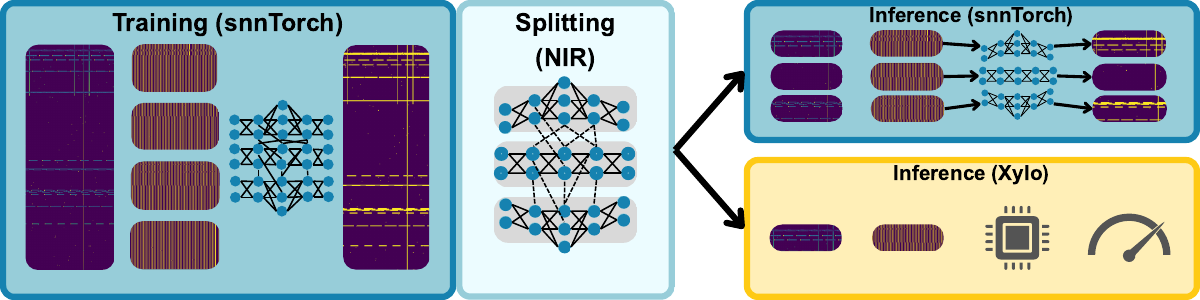}
    \caption[Overall methodology for RFI detection with spiking neural networks trained as a large single model and then split for inference on several neuromorphic chipsets.]{Overall methodology for RFI detection with spiking neural networks trained as a large single model and then split for inference on several neuromorphic chipsets. Spectrograms are split and latency encoded before feeding through the SNN, models are split in NIR format and deployed in snnTorch or to SynSense Xylo hardware for power measurement.}
    \label{fig:xylo:main-diagram}
\end{figure}

Our methodology, which Figure \ref{fig:xylo:main-diagram} outlines graphically, bridges the gap between high-level SNN training and deployment on resource-constrained neuromorphic hardware.
The core stages involve training a large SNN in snnTorch \cite{eshraghian_training_2023}, converting it to NIR format \cite{pedersen_neuromorphic_2024}, performing model splitting and adjustments in this format and deploying the models for inference either using snnTorch or SynSense Xylo hardware via the Rockpool library \cite{muir_rockpool_2019}.

The training process involves splitting spectrograms into model-width-sized patches.
Each patch is encoded into spikes using latency encoding and a set exposure time per spectrogram-timestep.
The resulting spike trains are input into a multi-layer feedforward SNN.
We calculate the loss and regularisation, and then backpropagate.
After training, the optimised SNN model is exported into NIR format and undergoes splitting using one of three algorithms (detailed later) to shard the model into smaller fixed-width models.
Following splitting, we handle these sub-modules in two ways.
Firstly, to evaluate accuracy degradation, we load all split sub-modules into snnTorch simultaneously for separate inference on spectrogram patches, allowing for comparison with the originally trained large model.
Secondly, for on-chip power evaluation, we load a single sub-model into Rockpool via NIR and deploy it onto real SynSense hardware, running a limited amount of inference to gain indicative power measurements.
\subsubsection{Model Architecture}
To simplify the model splitting process and ensure straightforward inference on the target hardware, we exclusively consider fully-connected feedforward SNN architectures in this work.
For a given input patch of width $W$, the network consists of an input layer with $W$ (spiking) neurons.
This is followed by an initial hidden layer of $W \cdot 2$ neurons, which then connects to $N-1$ subsequent hidden layers, each containing $H$ neurons. The final output layer mirrors the input, comprising $W$ neurons, to produce the RFI mask for each time step over $e$ exposure steps.
We conducted hyperparameter tuning with Optuna \cite{akiba_optuna_2019} optimising for F1-score including traditional hyper-parameters, snnTorch-specific decay rates ($\alpha$ and $\beta$), exposure-time, and fan-in regularisation weighting.

We expect high per-pixel accuracy results owing to the sparsity of RFI in the dataset, and therefore, we opted to optimise for F1-score.
Each hyper-parameter trial ran for 25 epochs with an initial learning rate of $1e-3$ while utilising Lightning's plateau scheduler, a fixed batch size of $36$ for all trials and 10\% of the original training set data.
Table \ref{tab:xylo:hyperparam-range} contains the ranges for each hyper-parameter tested; `Num Hidden' refers to the size of each hidden layer, `Num Layers' refers to the number of hidden layers, including a single scaling layer that is automatically set to be twice the network's input width and `Fan-in-weighting' refers to the weighting of the fan-in regularisation term.
Additionally, $\alpha$ and $\beta$ are shorthand methods to describe $\tau_{syn}$ and $\tau_{mem}$ representing $e^{(-\Delta t / \tau_{syn})}$ and $e^{(-\Delta t / \tau_{mem})}$ respectively.
All model input widths are tested over the entire parameter range, except the 8-channel input size models, which have a fixed hidden layer size of 64 neurons.
\begin{table}[!htbp]
\centering
\begin{tabular}{@{}cc@{}}
\toprule
Attribute  & Parameter Range \\ \midrule
Num Hidden & 128, 256, 512, 1024, 2048, 4096 \\
Num Layers & 2 - 6        \\
$\alpha$      & 0 - 1          \\
$\beta$       & 0 - 1    \\
Exposure   & 1 - 64         \\
Fan-in-weighting   & 0 - 0.1          \\ \bottomrule
\end{tabular}
\caption{Parameter ranges for attributes included in the final hyper-parameter searches.}
\label{tab:xylo:hyperparam-range}
\end{table}

Table \ref{tab:xylo:hera:optuna} contains the result of the hyper-parameter tuning.
We see that the 64-channel input models exhibit the best single-trial performance, followed by the smaller 8-channel and 32-channel models.
While performance by the larger model sizes may at first seem disappointing, the data volume used for hyper-parameter training and the lower epoch count likely leaves room for further performance compared to the smaller models since full-channel spectrograms are sliced into model-width-sized sub-spectrograms for training. Therefore, smaller models are trained for more iterations at a fixed number of epochs and data. 
Furthermore, we see that the smaller models make more use of the fan-in regularisation term, most model widths result in a scale difference between the $\alpha$ and $\beta$ decay parameters, and the 8-channel model in particular uses an exceptionally large exposure time and similarly scaled $\alpha$ and $\beta$ parameters for the spiking neurons.
\begin{table}[!htbp]
\centering
\begin{tabular}{cccccccc}
\hline
\begin{tabular}{@{}c@{}}Model\\Size\end{tabular} &\begin{tabular}{@{}c@{}}Num\\Layers\end{tabular}&\begin{tabular}{@{}c@{}}Num\\Hidden\end{tabular} & $\alpha$ & $\beta$  & Exposure &\begin{tabular}{@{}c@{}}Fan-in\\weighting\end{tabular}& F1       \\ \hline
8          & 4          & 64        & 0.127     & 0.261     & 62        & 0.042 & 0.945 \\
32         & 3          & 2048       & 0.187 & 0.073 & 14       & 0.063            & 0.938          \\
{\textbf{64}}   & 4          & 512        & 0.001 & 0.198 & 7        & 0.064            & \textbf{0.973} \\
128        & 3          & 512        & 0.15  & 0.06  & 11       & 0.074            & 0.887          \\
256        & 3          & 1024       & 0.166 & 0.414 & 21       & 0.009            & 0.818          \\
512        & 4          & 512        & 0.413 & 0.027 & 2        & 0.02             & 0.758     \\ \bottomrule   
\end{tabular}
\caption[Hyper-parameter search using Optuna multi-variate optimisation for the synthetic HERA dataset.]{Hyper-parameter search using Optuna multi-variate optimisation for the synthetic HERA dataset. Best F1-score and model size is in bold.}
\label{tab:xylo:hera:optuna}
\end{table}
Table \ref{tab:xylo:hera:optuna-params} contains final training parameters.
\begin{table}[!htb]
\centering
\begin{tabular}{ccccccc}
\hline
\begin{tabular}{@{}c@{}}Model\\Size\end{tabular} &\begin{tabular}{@{}c@{}}Num\\Layers\end{tabular}&\begin{tabular}{@{}c@{}}Num\\Hidden\end{tabular} & $\alpha$ & $\beta$  & Exp. &Fan-in\\ \hline
8          & 4          & 64        & 0.127     & 0.261     & 62        & 0.042  \\
32         & 3          & 2048       & 0.187 & 0.073 & 14       & 0.063                  \\
{\textbf{64}}   & 4          & 512        & 0.001 & 0.198 & 7        & 0.064            \\
128        & 3          & 512        & 0.15  & 0.06  & 11       & 0.074             \\
256        & 3          & 1024       & 0.166 & 0.414 & 21       & 0.009               \\
512        & 4          & 512        & 0.413 & 0.027 & 2        & 0.02              \\ \bottomrule   
\end{tabular}
\caption[Hyper-parameter search using Optuna multi-variate optimisation for the synthetic HERA dataset.]{Hyper-parameter search using Optuna multi-variate optimisation for the synthetic HERA dataset. We show the individual trials that performed best in the F1-score.}
\label{tab:xylo:hera:optuna-params}
\end{table}
\subsection{Model Splitting Algorithm}
\label{sec:xylo:methods:splitting}
\begin{figure}[!htb]
    \centering
    \includegraphics[width=1.0\linewidth]{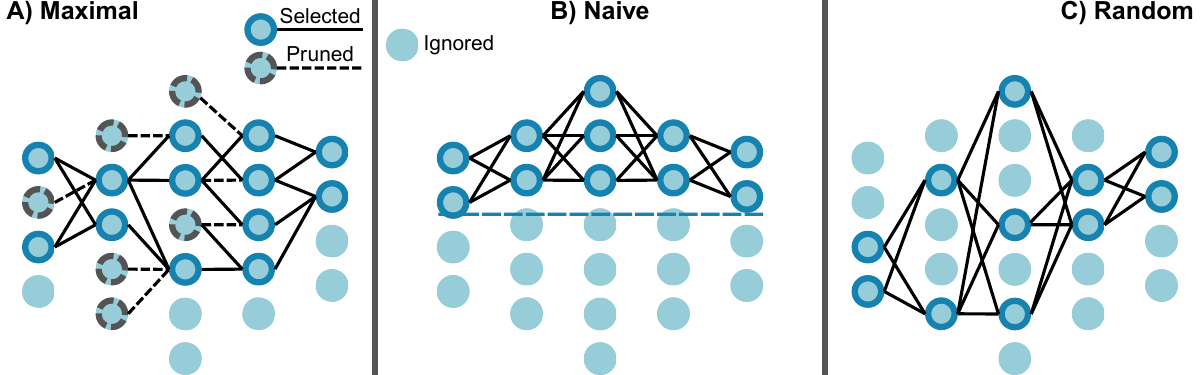}
    \caption[Model splitting approaches.]{Model splitting approaches. Our model splitting algorithm greedily selects a neuron bundle such that each neuron satisfies the hardware platform restrictions, then iteratively removes neurons until chipset restrictions are met. This is in contrast to a random or naive (geometric) splitting approach also depicted here.}
    \label{fig:xylo:model-splitting}
\end{figure}

\paragraph{Maximal Splitting} Maximal splitting takes in as input the original SNN model and a hardware configuration specifying the maximum input and output neurons, neuron fan-in ($f_{\text{in}}$), total number of hidden neurons, number of layers, and maximum total number of connections permitted by a particular hardware platform.
The algorithm's main intuition is to split the output neurons into suitably sized bundles, construct a sub-network that includes the $f_{\text{in}}$ most strongly connected inputs to each neuron in each layer, and then cull excess neurons iteratively, selecting the weakest connected neuron across all layers until fan-in, total neuron number and total connection number limits are met.
Figure \ref{fig:xylo:model-splitting}.a presents a graphical depiction of this approach, first selecting strongly connected neurons then culling greedily until hardware constraints are met.

Algorithm \ref{alg:xylo:illustrative_maximal_split} presents pseudocode for the maximal splitting algorithm.
For naive and random splitting strategies, the principal difference is in how $\verb|FindLeastImp|$ proceeds.
For a total neuron count $N$, the algorithm will process each neuron in each iteration and have no more than $N$ iterations providing a worst-case time complexity of $\mathcal{O}(N^2)$ complexity and $\mathcal{O}(N^2)$ memory requirements, to store each possible neuron connection in the worst-case allowing for arbitrary neuron connections between all layers.
In practice, however, the total number of iterations will be much lower than $N$ since at most $f_{\text{in}}\cdot N$ neurons will be included in the initial maximal graph and typically $f_\text{in} << N$.
\paragraph{Naive Splitting}
Naive splitting works by setting both the input and output neuron bundles simultaneously; naively selecting adjacent $m_{\text{width}}$ input and output neurons for each split.
The culling process is driven by selecting the geometrically furthest neuron from the input and output neuron, resulting in a tight neuron bundle from input to output depicted in Figure \ref{fig:xylo:model-splitting}.b.
\paragraph{Random Splitting}
Finally, random splitting selects a random initial sub-network, then randomly culls neurons (including input connections) from each layer in a weighted random choice (weighted by the number of neurons in each layer).
The resulting sub-network can exhibit an expected random structure of disparate neurons depicted in Figure \ref{fig:xylo:model-splitting}.c.
\begin{algorithm}[t!]
\caption{Maximal SNN Splitting}\label{alg:xylo:illustrative_maximal_split}
\begin{algorithmic}[1]
\STATE \textbf{Procedure} MaximalSplit($m_{\text{orig}}$, $C_{\text{hw}}$)
    \STATE $\text{output} \leftarrow []$
    \STATE $\text{allConns} \leftarrow \text{AllPossConns}(m_{\text{orig}})$
    \STATE $\text{outBundles} \leftarrow \text{SplitOutputs}(m_{\text{orig}}, C_{\text{hw}}\text{.width})$
    \FOR{$\text{outB}$ in $\text{outBundles}$}
        \STATE $\text{currCs} \leftarrow \text{allConns.copy}$
        
        \COMMENT{Phase 1: Prune Hidden}
        \WHILE{$\text{HN}(\text{currCs}) > C_{\text{hw}}.\text{maxN}$}
            \STATE $\text{nImps} \leftarrow \text{CalcImp}(\text{currCs}, \text{outB}, \text{`hidden'})$
            \STATE $\text{leastImpNeuron} \leftarrow \text{FindLeastImp}(\text{nImps})$
            \STATE $\text{currCs} \leftarrow \text{RemBackwards}(\text{leastImpNeuron})$
        \ENDWHILE

        \COMMENT{Phase 2: Prune Inputs}
        \WHILE{$\text{InNeurons}(\text{currCs}) > C_{\text{hw}}.\text{maxInNeurons}$}
            \STATE $\text{nImps} \leftarrow \text{CalcImp}(\text{currCs}, \text{outB}, \text{`input'})$ 
            \STATE $\text{leastImpIn} \leftarrow \text{FindLeastImp}(\text{nImps})$
            \STATE $\text{currCs} \leftarrow \text{RemForwards}(\text{leastImpIn})$
        \ENDWHILE
        
        \COMMENT{Phase 3: Prune Outputs}
        \WHILE{$\text{OutConns}(\text{currCs}) > C_{\text{hw}}.\text{maxOutFIn}$}
             \STATE $\text{nImps} \leftarrow \text{CalcImp}(\text{currCs}, \text{outB}, \text{`output'})$
             \STATE $\text{leastImpLast} \leftarrow \text{FindLeastImp}(\text{nImps})$
             \STATE $\text{currCs} \leftarrow \text{RemBackwards}(\text{leastImpLast})$ 
        \ENDWHILE

        \COMMENT{Phase 4: Prune Total Connections}
        \WHILE{$\text{TConns}(\text{currCs}) > C_{\text{hw}}.\text{maxConns}$}
            \STATE $\text{nImps} \leftarrow \text{CalcImp}(\text{currCs}, \text{outB}, \text{`hidden'})$
            \STATE $\text{leastImpNeuron} \leftarrow \text{FindLeastImp}(\text{nImps})$
            \STATE $\text{currCs} \leftarrow \text{RemBackwards}(\text{leastImpNeuron})$
        \ENDWHILE
        
        \STATE $\text{output.append}(\text{BuildSubModel}(\text{currCs}))$
    \ENDFOR
    \RETURN $\text{output}$
\end{algorithmic}
\end{algorithm}
While straightforward, these model splitting algorithms deliver hardware-compatible sub-graphs suitable for inference on real neuromorphic hardware and represent a first step towards high-level compilation of large SNNs to numerous neuromorphic chipsets.
\section{Experiments}\label{sec:xylo:experiments}
We evaluate our model through several experiments with the Hydrogen Epoch of Reionisation Array (HERA) dataset \cite{mesarcik_learning_2022}.
First, by training full-sized models of varying input widths.
Second, we provide energy and power estimates based on measuring indicative spike-rates of each arbitrary-width model.
Third, we split our full-width models into Xylo-compatible sub-modules using our proposed model splitting algorithms and report on detection accuracy.
Fourth, we provide power usage measurements for split models on real Xylo hardware.
Fifth, we ablate our fan-in regularisation technique, and, finally, we compare model performance against the SoTA in algorithmic, ANN, and SNN-based RFI detection methods, finding our full-width models achieve SoTA detection performance among SNN baselines.
%We report per-pixel accuracy, AUROC, AUPRC, and F1-score, focusing our analysis on the class-balanced AUPRC and F1-score metrics, in line with prior literature, reflecting the relative sparsity of RFI over a typical observation.
%We conducted all training on 8 AMD MI250X GPUs, for 50 epochs per trial with a batch size of 36.
%We trained with a learning-rate of $1E-3$ while using Lightning's plateau scheduler and Adam as the optimiser.
\subsection{Training Environment}\label{app:xylo:experiments:env}
All experiments were run on single HPE Cray EX nodes featuring an AMD EPYC 7A53 `Trento' 64-core CPU and eight AMD Instinct MI250X GPUs.
Original models were trained with snnTorch 0.9.4 \cite{eshraghian_training_2023} using PyTorch 1.13.1+ROCm5.2 and Lightning 2.2.0 \cite{falcon_pytorch_2019}.
Conversion to Xylo-compatible models was completed with Rockpool 2.9.1 \cite{muir_rockpool_2019} and NIR 1.0.4 \cite{pedersen_neuromorphic_2024}.
We patched Rokpool's NIR reading module to fix the $dt$ value to $1E-4$ for CUBA-LiF neurons, as this value is inferred from the $\tau_{mem}$ and $\tau_{syn}$ values written out to NIR format by default.
We trained all models in full precision for 50 epochs using mini-batches of 36 samples, using Adam as the optimiser and a learning rate of $1E-3$.
We utilised Lightning's reduce-lr-on-plateau scheduler (factor 0.5, patience 10) to govern learning-rate decay.
Training-validation splits were deterministic, training data were shuffled each epoch, and testing data order was fixed.
Hyper-parameter tuning trials take less than six hours each, full training runs less than 24 each, and all subsequent model splitting, power estimate, and power measurement trials take approximately 15 minutes each.
Distributed data-parallel execution spanned all eight GPUs for training, but NIR conversion occurred on a single GPU for final testing.
The final model corresponds to the last saved checkpoint.
Finally, for power measurement, we utilised a SynSense Xylo Audio 2 development kit for power consumption measurement \cite{synsense_ag_xylo_2022} paired with the Samna 0.45.3 runtime environment.
\subsection{Results on HERA Dataset}
\begin{figure}[!htbp]
    \centering
    \begin{subfigure}{0.49\columnwidth}
        \centering
        \includegraphics[height=1.5in, keepaspectratio]{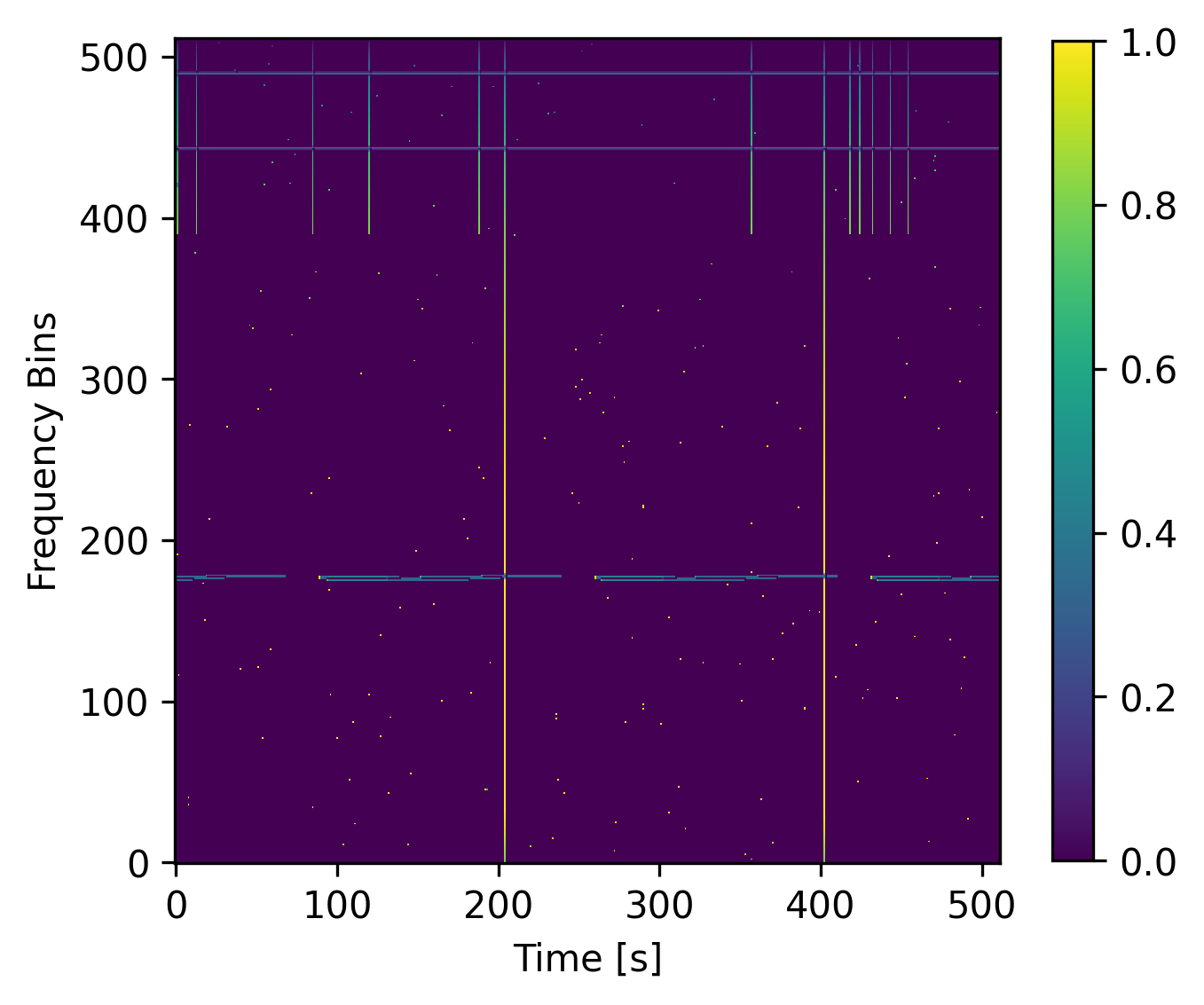}
        \caption{Original spectrogram.}
        \label{fig:xylo:res:hera:orig}
    \end{subfigure}
    \begin{subfigure}{0.49\columnwidth}
        \centering
        \includegraphics[height=1.5in, keepaspectratio]{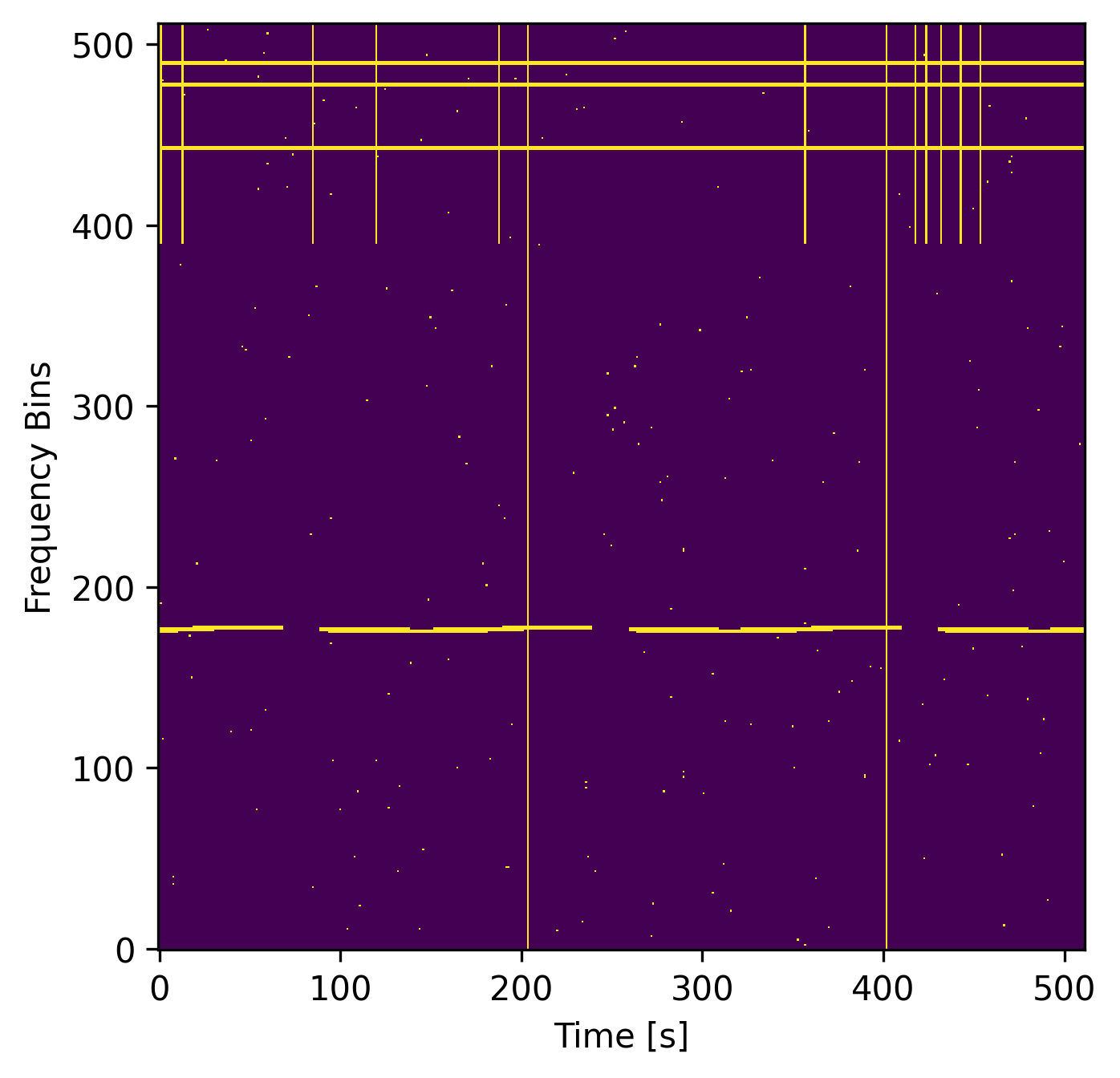}
        \caption{Ground-truth mask.}
        \label{fig:xylo:res:hera:mask}
    \end{subfigure}
    \caption{An original full-size spectrogram and its corresponding ground-truth label.}
\label{fig:xylo:example:orig}
\end{figure}

\begin{figure}[!htbp]
    \centering
    \begin{subfigure}{0.49\columnwidth}
        \centering
        \includegraphics[height=1.5in, keepaspectratio]{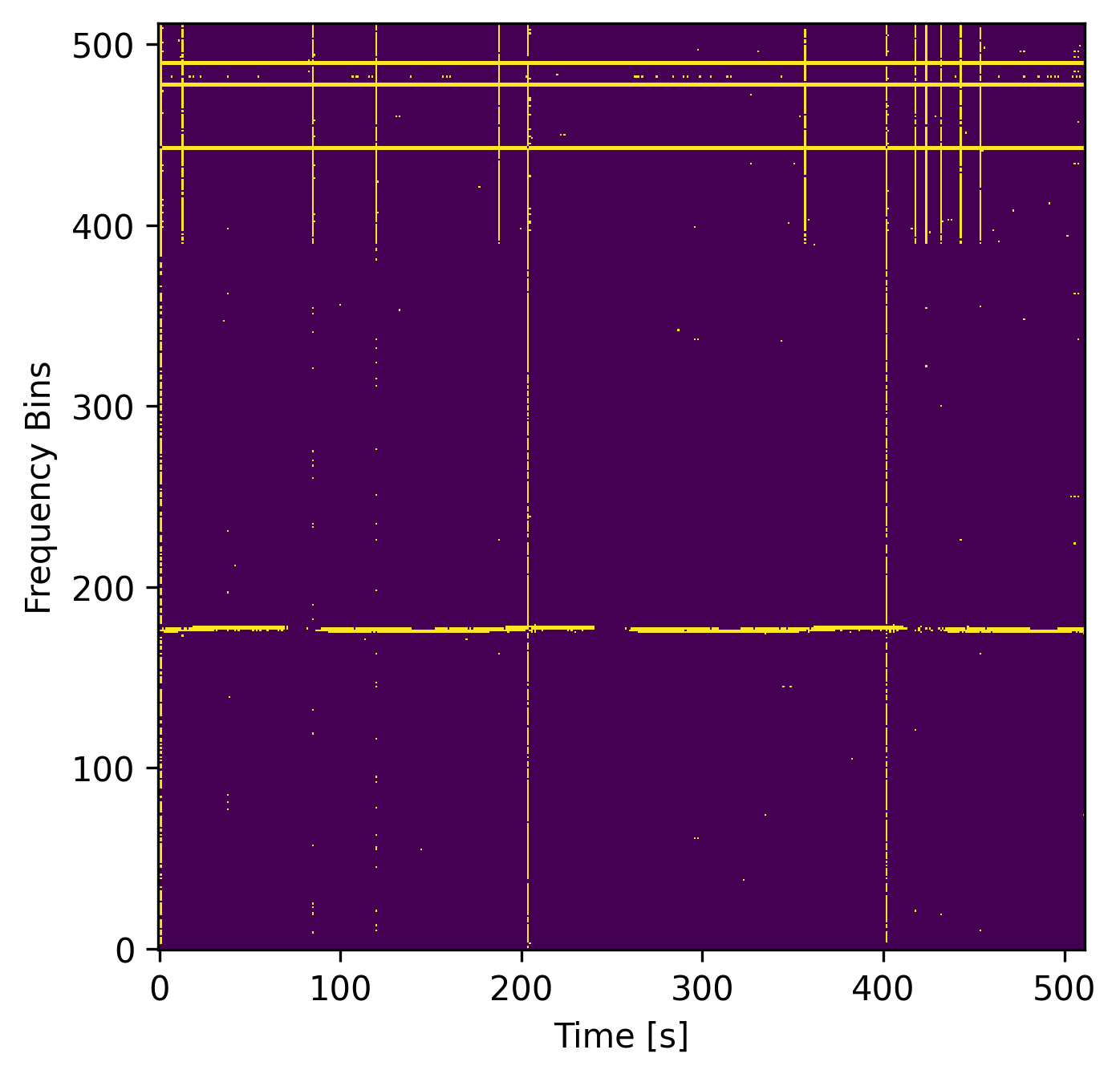}
        \caption{512 channel inference.}
        \label{fig:xylo:res:512:full}
    \end{subfigure}
    \begin{subfigure}{0.49\columnwidth}
        \centering
        \includegraphics[height=1.5in, keepaspectratio]{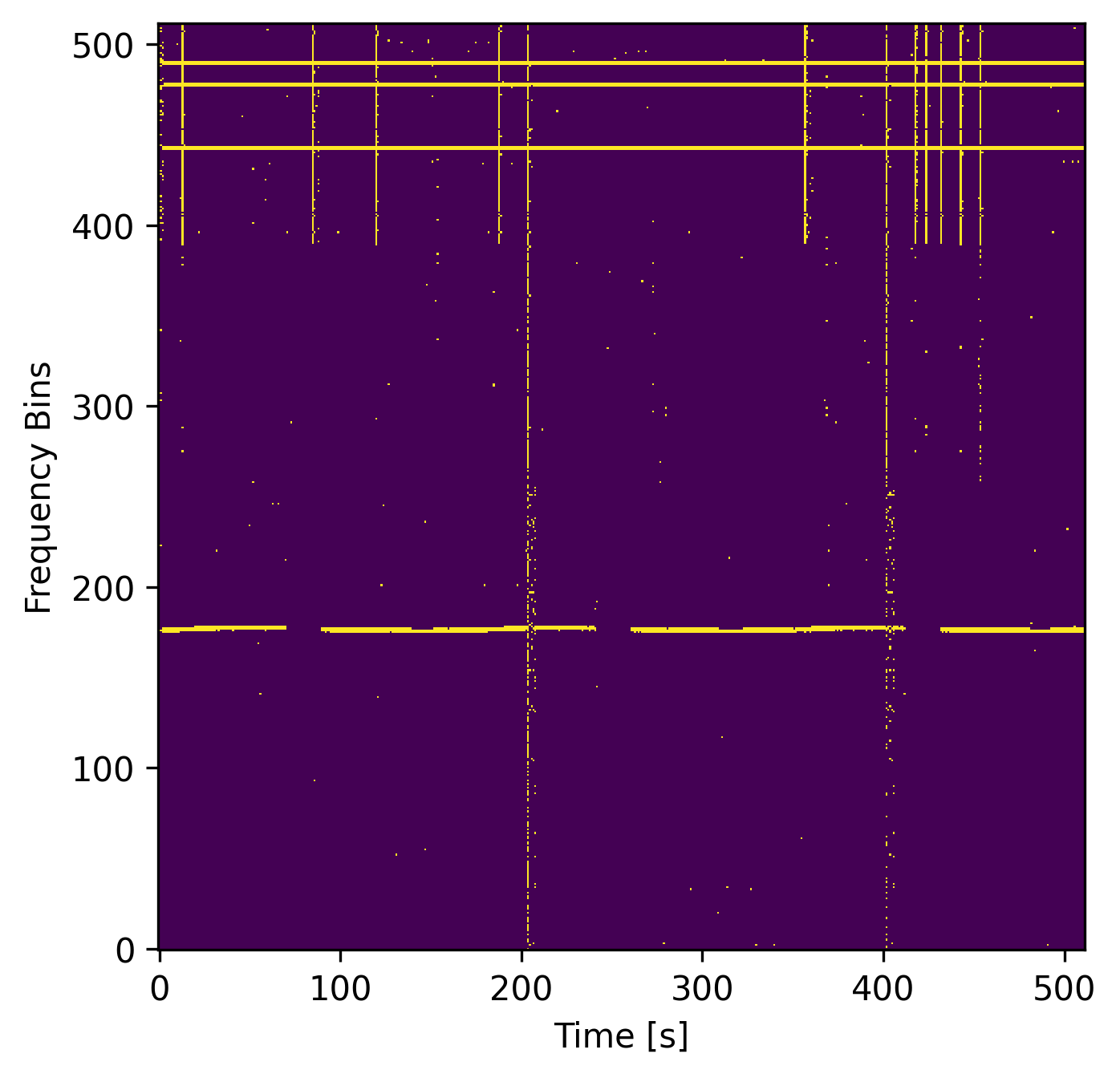}
        \caption{256 channel inference.}
        \label{fig:xylo:res:256:full}
    \end{subfigure}
    \begin{subfigure}{0.49\columnwidth}
        \centering
        \includegraphics[height=1.5in, keepaspectratio]{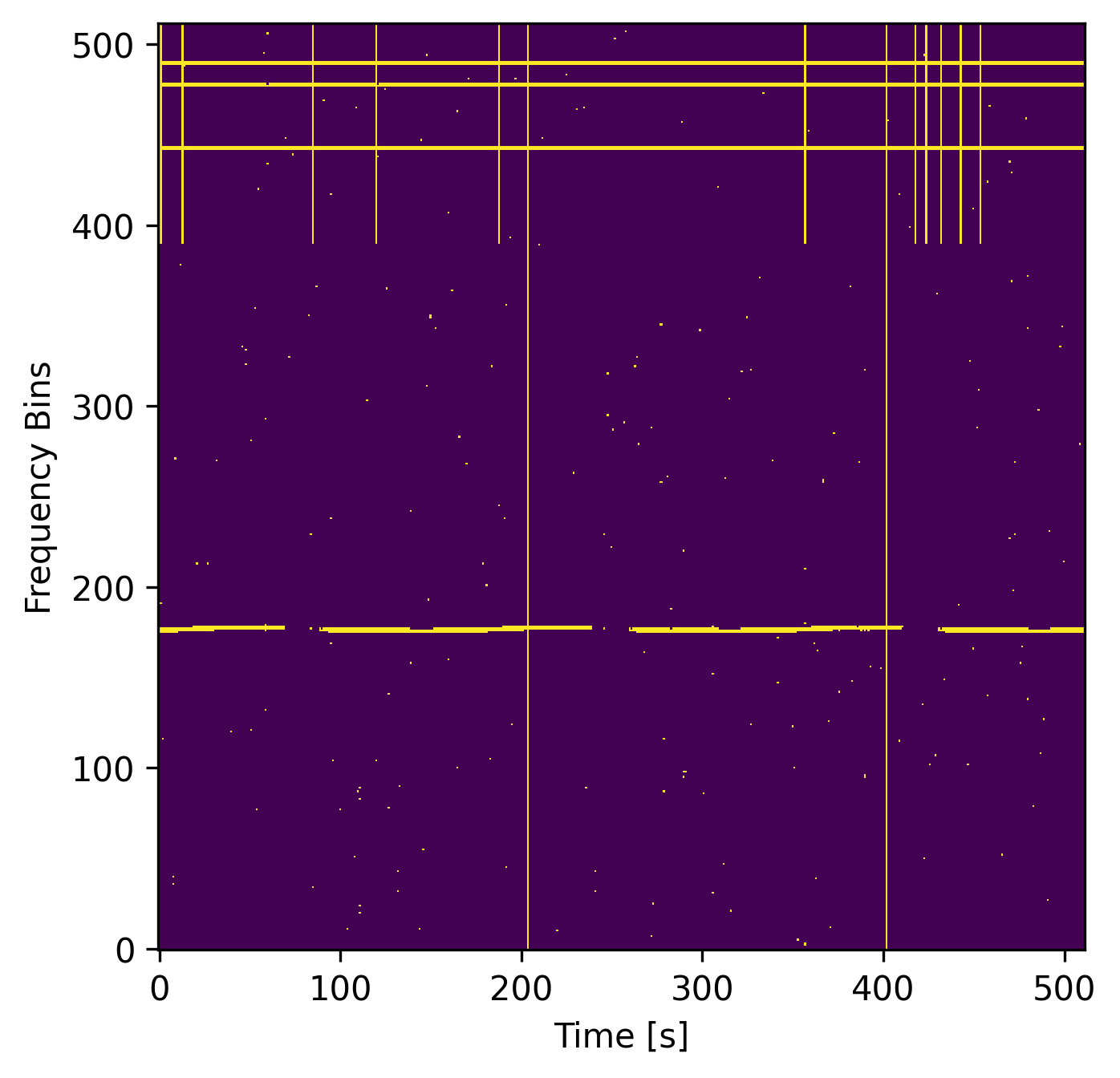}
        \caption{128 channel inference.}
        \label{fig:xylo:res:128:full}
    \end{subfigure}
    \begin{subfigure}{0.49\columnwidth}
        \centering
        \includegraphics[height=1.5in, keepaspectratio]{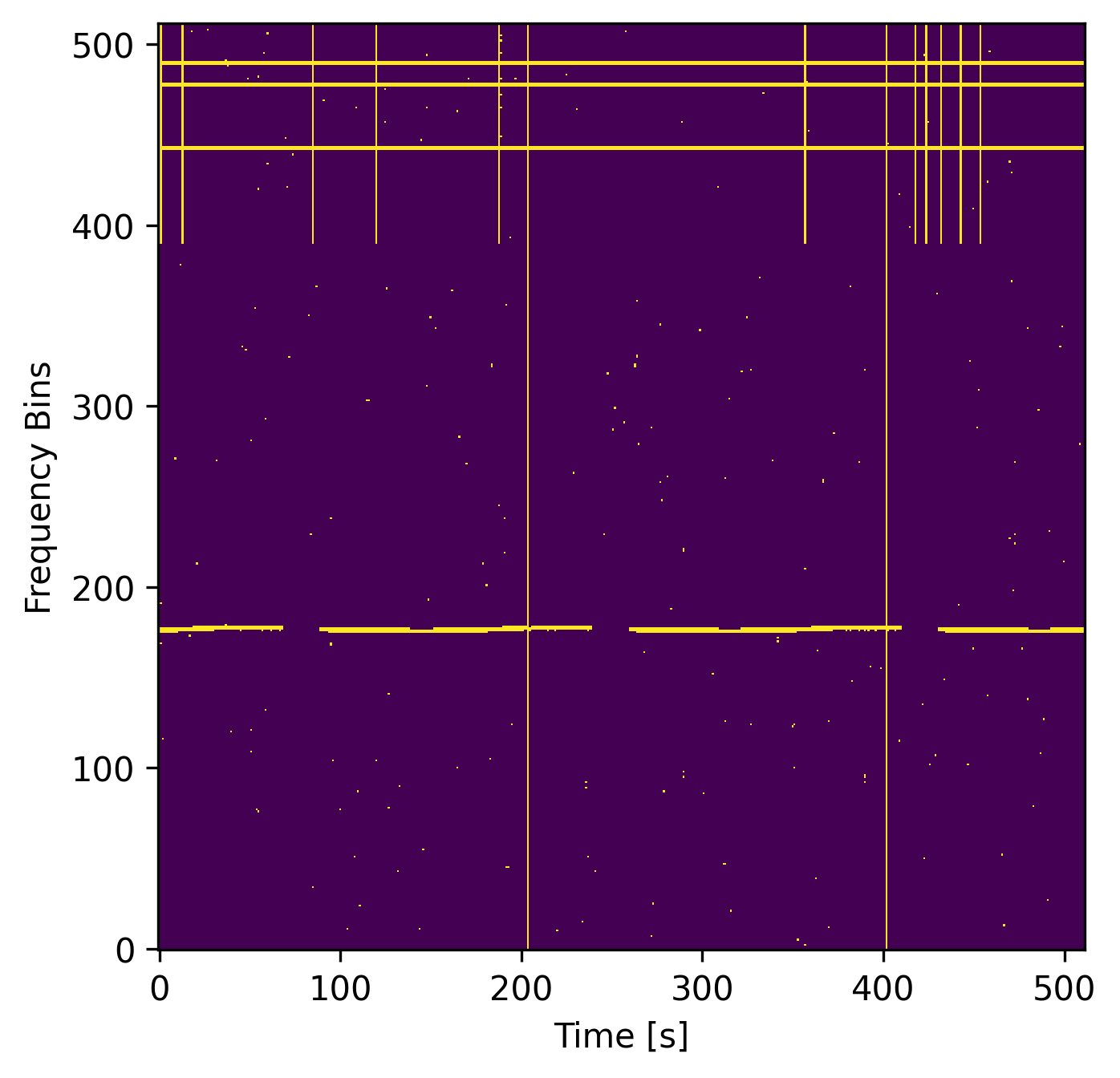}
        \caption{64 channel inference.}
        \label{fig:xylo:res:64:full}
    \end{subfigure}
    \begin{subfigure}{0.49\columnwidth}
        \centering
        \includegraphics[height=1.5in, keepaspectratio]{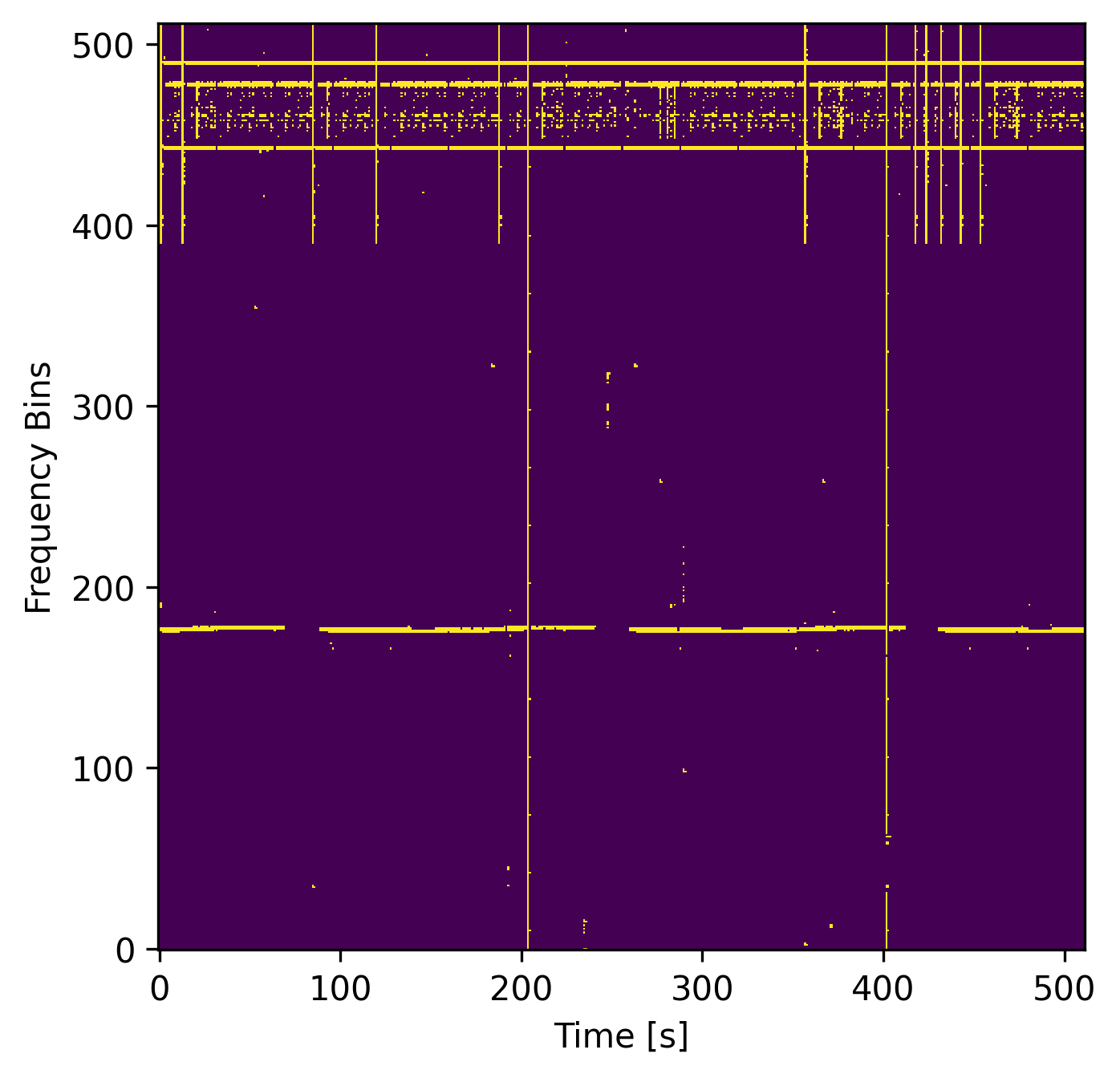}
        \caption{32 channel inference.}
        \label{fig:xylo:res:32:full}
    \end{subfigure}
    \begin{subfigure}{0.49\columnwidth}
        \centering
        \includegraphics[height=1.5in, keepaspectratio]{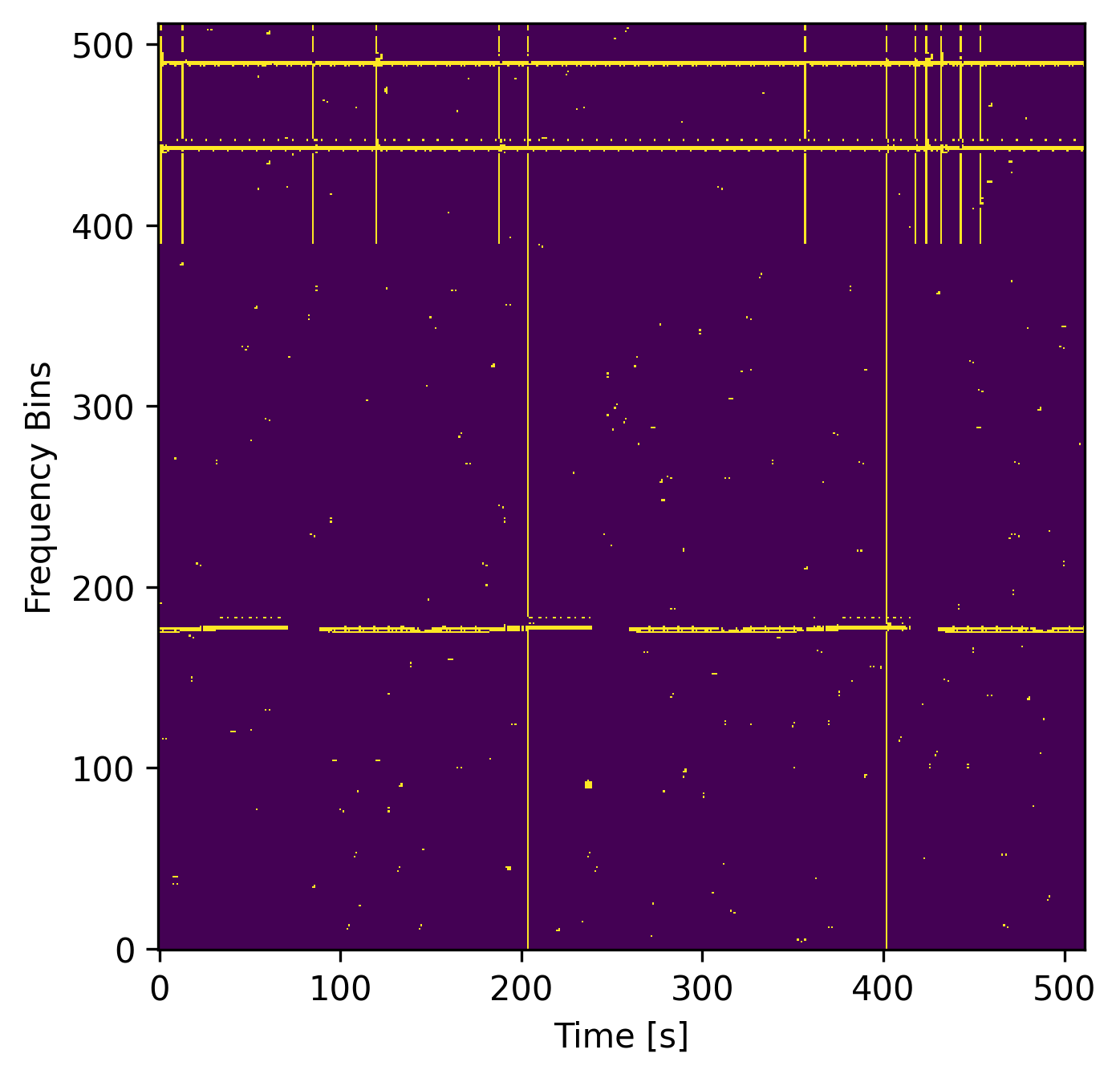}
        \caption{8 channel inference.}
        \label{fig:xylo:res:8:full}
    \end{subfigure}
    \caption[Inference examples across different model sizes.]{Inference examples across different model sizes. Examples of the original spectrogram in Figure \ref{fig:xylo:example:orig} with varying trained model widths and their corresponding hardware reconstructions. We see that in all widths, hardware faithfully reconstructs the output of the software-trained models but that smaller models outperform larger input widths.}
\label{fig:xylo:example:hera}
\end{figure}

First, we train models of arbitrary input width as baselines to split into sub-modules.
Table \ref{tab:xylo:hera:full} contains detection performance results averaged over 10 trials.
We observe diminishing returns in accuracy and AUROC beyond 128 input channels, and our 64-channel model achieves the best performance in class-balanced metrics, AUPRC, and F1-score.
We also train an 8-channel model suitable for direct deployment on Xylo hardware, providing a baseline for comparison with model splitting techniques.
\begin{table}[!htb]
\centering
\begin{tabular}{@{}ccccccccc@{}}
\toprule
Model Size & \multicolumn{2}{c}{Accuracy}& \multicolumn{2}{c}{AUROC}& \multicolumn{2}{c}{AUPRC}& \multicolumn{2}{c}{F1}\\ \midrule
8          & 0.9903         &0.001& 0.7501       &0.019& 0.8908       &0.024& 0.8568     &0.032\\
32         & 0.9906         &0.002& 0.9321       &0.010& 0.8834       &0.018& 0.8771     &0.019\\
64         & 0.9986         &0.000& 0.9880       &0.001& \textbf{0.9833} &0.002& \textbf{0.9827}  &0.002\\
128        & \textbf{0.9996}  &0.000& \textbf{0.9883}     &0.003& 0.9780  &0.006& 0.9777 &0.006\\
256        & 0.9912         &0.001& 0.9133  &0.012& 0.8541  &0.017& 0.8507  &0.018\\
512        & 0.9897         &0.002& 0.9311  &0.014& 0.8746  &0.024& 0.8718   &0.025\\ \bottomrule
\end{tabular}
\caption[Detection performance results for full-sized hardware-regularised models of varying sizes on the HERA dataset.]{Detection performance results for full-sized hardware-regularised models of varying sizes on the HERA dataset. Scores are presented as mean and standard deviation over 10 trials, and the best scores are bolded. We see larger, but not the largest models make best use of the available data and perform significantly better than larger or smaller data widths.}\label{tab:xylo:hera:full}
\end{table}
Figure \ref{fig:xylo:example:orig} presents an original spectrogram (Figure \ref{fig:xylo:res:hera:orig}), and the associated ground-truth mask (Figure \ref{fig:xylo:res:hera:mask}), and Figure \ref{fig:xylo:example:hera} presents example inference for full-width models at varying input widths.
We can see that while all models detect the major RFI features, smaller models struggle at the boundaries and produce noisier outputs especially visible in the 32-channel example, Figure \ref{fig:xylo:res:32:full}.

\FloatBarrier
\subsection{Power Consumption Estimates}
Next, we prepare power consumption estimates for our full-width models by running each trained model through the test set, storing spike counts for each layer using a method similar to prior literature \cite{barchid_spiking_2023}.
We estimate the synaptic operations for each layer $l$ from the following:
\begin{equation}
    SOPs(l) = C_{in} \times C_{out} \times R_s(l)
\end{equation}
where $C_{in}$ and $C_{out}$ are the input and output sizes of a particular layer and $R_s(l)$ is the spike rate for layer $l$.
We then provide an estimate by considering the spike-based accumulation (AC) operations utilising 40nm CMOS technology at $E_{AC} = 0.9$pJ \cite{horowitz_11_2014}.
We use the measured spike-rates of each layer combined with power estimates for accumulate operations to provide a per-spectrogram energy usage estimate, which we can then translate into power measurements at 50MHz (modelling typical Xylo deployment speeds) and also at the minimum clock frequency required for real-time operation in the HERA telescope (an integration time of 3.52 seconds).
These values provide a theoretical energy estimate for inference, which we then normalise, factoring in various options for compute time and the exposure time of each model-width's hyperparameters.
We calculate power consumption estimates by first estimating the inference time of a single spectrogram by dividing the number of SOPs by the clockspeed in question.
Then, we divide the energy usage estimate for a spectrogram by this value to provide a power estimate given the expected inference time over a single spectrogram.

We multiply the baseline estimate for a model of a particular width to estimate consumption at the full spectrogram channel width (512).
Hypothetical real-time frequencies are intentionally set at extremely low levels to provide a conservative lower estimate.
Table \ref{tab:xylo:energy-estimate} contains energy and power measurements for each full-size model.
Intriguingly, despite the different model scales, the differences in exposure times roughly equalise (except for 32 and 256 channel models) the eventual synaptic operation count, and therefore the overall power consumption.
Based on modelling alone, which does not account for device-specific overheads, we expect power usage to range from $<100$mW using the default clock rate down to or below 1 mW at extremely low clock speeds.
\begin{table}[!htb]
\centering
\begin{tabular}{@{}ccccccc@{}}
\toprule
\begin{tabular}{@{}c@{}}Model\\Size\end{tabular} & Exp & \begin{tabular}{@{}c@{}}Real-Time\\Hz\end{tabular}& SOPs     & \begin{tabular}[c]{@{}c@{}}SE\\(mJ)\end{tabular} & \begin{tabular}[c]{@{}c@{}}SPE at\\ 50MHz\\(mW)\end{tabular} & \begin{tabular}[c]{@{}c@{}}SPE at\\RTF\\(mW)\end{tabular} \\ \midrule
8          & 62       & 17.61        & 3.66E+7 & 0.508 & 44.4                                                                                               & 1.80E-2                                                                                    \\
32         & 14       & 3.98         & 9.69E+8 & 11.9                                      & 9.79                                                                                              & 1.05E-1                                                                                    \\
64         & 7        & 1.99         & 3.38E+7 & \textbf{0.419}                                     & \textbf{5.08}                                                                                             & 1.86E-3                                                                                \\
128        & 11       & 3.13         & 1.73E+7 & 0.640                                    & 7.39                                                                                            & 1.42E-3                                                                                 \\
256        & 21       & 5.97         & 6.61E+7 & 60.4                                     & 91.5                                                                                         & 6.71E-2                                                                               \\
512        & 2        & 0.568        & 2.10E+6 & 0.778                                  & 18.5                                                                                              & \textbf{4.32E-4}                                                                             \\ \bottomrule
\end{tabular}
\caption[Energy and power estimates for SNN-based RFI detection.]{Energy and power estimates for SNN-based RFI detection. The table details energy per spectrogram (SE) and power estimate (SPE) across various model sizes operating at 50 MHz and a Real-Time Frequency (RTF) dependent on the exposure time used. Results presented as mean over 10 trials and best values are bolded.}\label{tab:xylo:energy-estimate}
\end{table}
\subsection{Effect of Model Splitting}
We took each trained full-width model and split each into Xylo-compatible sub-modules using one of three splitting algorithms outlined earlier.
Table \ref{tab:xylo:hera:splitting} contains results for running all sub-models together to detect RFI over entire spectrograms.
We observe that while per-pixel accuracy is not significantly degraded, the performance degradation for AUPRC and F1 Scores is substantial.
We also observe that for these regularised models, the best-performing splitting algorithm is maximal splitting up to 128-channel models, indicating that the fan-in limits for each neuron impose an upper limit on the maximum size of the original, unsplit model.
\begin{table}[htb!]
\centering
\begin{tabular}{@{}cccccccccc@{}}
\toprule
\begin{tabular}{@{}c@{}}Model\\Size\end{tabular}           & \begin{tabular}{@{}c@{}}Splitting\\Method\end{tabular} & \multicolumn{2}{c}{{Accuracy}}& \multicolumn{2}{c}{{AUROC}}& \multicolumn{2}{c}{{AUPRC}}& \multicolumn{2}{c}{{F1}}\\ \midrule
\multirow{3}{*}{32}  & Maximal          & 0.9328        &0.013& \textbf{0.5944}       &0.054& 0.2139       &0.084& 0.1893      &0.066\\
                     & Naive            & 0.9437              &0.008& 0.5195            &0.029& 0.1044            &0.061& 0.0974          &0.029\\
                     & Random           & 0.9247         &0.011& 0.5513           &0.041& 0.1423            &0.062& 0.1288          &0.043\\
\multirow{3}{*}{64}  & Maximal          & 0.9471              &0.009& 0.5487            &0.042& 0.2720           &0.111& 0.1701      &0.084\\
                     & Naive            & 0.9583               &0.002& 0.5363           &0.015& 0.3653          &0.065& 0.1496         &0.033\\
                     & Random           & 0.9548               &0.005& 0.5353             &0.035& 0.4004            &0.072& 0.1529         &0.085\\
\multirow{3}{*}{128} & Maximal          & 0.9583               &0.004& 0.5655             &0.030& 0.3167           &0.097& \textbf{0.2081}        &0.078\\
                     & Naive            & 0.9577               &0.002& 0.5185            &0.008& 0.3136            &0.042& 0.0951        &0.016\\
                     & Random           & 0.9552             &0.004& 0.5252             &0.019& 0.2455             &0.100& 0.1126        &0.042\\
\multirow{3}{*}{256} & Maximal          & 0.5367             &0.215& 0.3878          &0.066& 0.1402        &0.055& 0.0786         &0.002\\
                     & Naive            & 0.8684             &0.037& 0.4953           &0.022& 0.0803         &0.039& 0.0795      &0.007\\
                     & Random           & 0.8888            &0.025& 0.4992            &0.014& 0.0757          &0.025& 0.0789       &0.003\\
\multirow{3}{*}{512} & Maximal          & 0.9507               &0.010& 0.5021           &0.006& 0.2448         &0.211& 0.0786        &0.002\\
                     & Naive            & \textbf{0.9603}               &0.001& 0.5011           &0.001& 0.4753        &0.059& 0.0765        &0.002\\
                     & Random           & 0.9591              &0.001& 0.5010           &0.002& \textbf{0.5122}           &0.024& 0.0787         &0.002\\ \bottomrule 
\end{tabular}
\caption[Detection performance results for regularised split models on the HERA dataset.]{Detection performance results for regularised split models on the HERA dataset. Scores presented as mean and standard deviation over 10 trials and best scores are bolded. Maximal splitting performs best followed by naive and then random approaches for 32, 64 and 128 width models while splitting introduces irrecoverable error for larger model widths.}\label{tab:xylo:hera:splitting}
\end{table}
\subsection{Power Consumption Measurements}
For each full model width, we split 10 regularised models with maximal splitting and measure power consumption over approximately five seconds of inference.
We provide measurements at both the default 50MHz clock speed supported by the Xylo 2 board and the minimal supported clock speed of 6.25 MHz. 
The Xylo HDK provides tooling to measure power consumption, isolating the SNN core itself.
Table \ref{tab:xylo:energy-measured} contains power consumption measurements for varying model sizes.
We multiply all measurements and estimates for a single chipset by $\frac{512}{8} = 64$ times to provide values on a per-spectrogram basis.
We find that power consumption is extremely low across all splittings, validating our observation of roughly equal Synaptic operations in Table \ref{tab:xylo:energy-estimate}.

The spectrogram-scaled power consumption is of particular note, representing the power consumption required to detect RFI in one of several thousand baselines.
%HERA has 350 antennae and 61,075 baselines with 512 inference channels each, and SKA-Low requires processing of more than 150 thousand baselines with 208,333 frequency channels each \cite{vermij_challenges_2015}.
While a more comprehensive investigation is justified, the potential efficiency gains are immense.
\begin{table}[!htb]
\centering
\begin{tabular}{@{}cccccccccc@{}}
\toprule
\begin{tabular}{@{}c@{}}Original\\Model\\Size\end{tabular} & Exposure & \multicolumn{2}{c}{\begin{tabular}[c]{@{}c@{}}MP at\\50MHz\\(mW)\end{tabular}}& \multicolumn{2}{c}{\begin{tabular}[c]{@{}c@{}}SP at\\50MHz\\(mw)\end{tabular}}& \multicolumn{2}{c}{\begin{tabular}[c]{@{}c@{}}MP at\\6.25MHz\\(mW)\end{tabular}}& \multicolumn{2}{c}{\begin{tabular}[c]{@{}c@{}}SP at\\6.25MHz\\(mW)\end{tabular}}\\ \midrule
8                   & 62       & 1.63                                                                            &0.205& 104.5                                                                                    &13.1& 0.224                                                                            &0.001& 14.3                                                                                       &0.072\\
32                  & 14       & 1.28                                                                             &0.053& 81.7                                                                                     &3.37& 0.239                                                                            &0.007& 15.3                                                                                      &0.461\\
64                  & 7        & 1.22                                                                             &0.033& 78.1                                                                                 &2.13& 0.235                                                                         &0.007& 15.0                                                                                    &0.455\\
128                 & 11       & 1.23                                                                          &0.041& 78.4                                                                                  &2.64& 0.234                                                                           &0.004& 15.0                                                                                     &0.288\\
256                 & 21       & 1.59                                                                             &0.025& 102                                                                                     &1.59& 0.237                                                                            &0.006& 15.1                                                                                      &0.359\\
512                 & 2        & 1.20                                                                            &0.019& 76.8                                                                                     &1.22& 0.239                                                                             &0.003& 15.3                                                                                       &0.184\\ \bottomrule
\end{tabular}
\caption[Energy and power measurements for SNN-based RFI detection measured on SynSense Xylo Hardware.]{Energy and power measurements for SNN-based RFI detection measured on SynSense Xylo Hardware. Contains model-power (MP) and spectrogram-power (SP) for a single sub-module in varying split sizes.} Results are presented as mean and standard deviation over 10 trials.\label{tab:xylo:energy-measured}
\end{table}
Table \ref{tab:xylo:energy-measured:full} presents power consumption measurements for all three model splitting algorithms averaged over available trials.
%Here we present power consumption measurements for all three model splitting algorithms averaged over available trials in Table \ref{tab:energy-measured:full}.
We see that power usage is similar regardless of the algorithm used, indicating a consistent and high usage of available resources.
We also note the consistency in power usage across all models and splitting algorithms.
\begin{table}[!htb]
\centering
\begin{tabular}{@{}ccccccc@{}}
\toprule
\begin{tabular}{@{}c@{}}Model\\Size\end{tabular} & Exposure            & \begin{tabular}{@{}c@{}}Splitting\\Algorithm\end{tabular} & \multicolumn{2}{c}{\begin{tabular}[c]{@{}c@{}}Model Power\\at 50MHz\\(mW)\end{tabular}}& \multicolumn{2}{c}{\begin{tabular}[c]{@{}c@{}}Model Power\\at 6.25MHz\\(mW)\end{tabular}}\\ \midrule
\multirow{3}{*}{8}    & \multirow{3}{*}{62} & Maximal             & 1.63 &0.205& 0.22
 &0.001\\
                      &                     & Naive               & 1.63 &0.201& 0.22
 &0.002\\
                      &                     & Random              & 1.63 &0.202& 0.22
 &0.002\\
\multirow{3}{*}{32}   & \multirow{3}{*}{14} & Maximal             & 1.28 &0.053& 0.24
 &0.007\\
                      &                     & Naive               & 1.28 &0.044& 0.23
 &0.006\\
                      &                     & Random              & 1.29 &0.038& 0.24
 &0.005\\
\multirow{3}{*}{64}   & \multirow{3}{*}{7}  & Maximal             & 1.22 &0.033& 0.23
 &0.007\\
                      &                     & Naive               & 1.23 &0.043& 0.23
 &0.007\\
                      &                     & Random              & 1.22 &0.026& 0.24
 &0.009\\
\multirow{3}{*}{128}  & \multirow{3}{*}{11} & Maximal             & 1.23 &0.041& 0.23 &0.004\\
                      &                     & Naive               & 1.22 &0.030& 0.24
 &0.010\\
                      &                     & Random              & 1.20 &0.027& 0.23
 &0.009\\
\multirow{3}{*}{256}  & \multirow{3}{*}{21} & Maximal             & 1.59 &0.025& 0.24
 &0.006\\
                      &                     & Naive               & 1.59 &0.025& 0.24
 &0.007\\
                      &                     & Random              & 1.56 &0.019& 0.24
 &0.004\\
\multirow{3}{*}{512}  & \multirow{3}{*}{2}  & Maximal             & 1.20 &0.019& 0.24
 &0.003\\
                      &                     & Naive               & 1.22 &0.004& 0.23
 &0.009\\
                      &                     & Random              & 1.19 &0.055& 0.24 &0.005\\ \bottomrule
\end{tabular}
\caption[Energy and power measurements for SNN-based RFI detection measured on SynSense Xylo Hardware with varying splitting algorithms.]{Energy and power measurements for SNN-based RFI detection measured on SynSense Xylo Hardware with varying splitting algorithms. The table details power usage for a single sub-module in varying split sizes operating at 50MHz and 6.25 MHz. Measurements are presented as mean and standard deviation over 10 trials.}
\label{tab:xylo:energy-measured:full}
\end{table}
\subsection{Ablation Study: Hardware-Aware Regularisation}
Table \ref{tab:xylo:ablate:hw} compares the best-performing splitting model algorithm for full models with and without hardware-aware regularisation.
While improvements are modest, the difference is more significant for the best-performing 64 and 128 models.
We also note that the best-performing splitting algorithm is the maximal algorithm up until 128 channel models.
This suggests an upper limit imposed by hardware restrictions as the learned dynamics in the larger networks outscale what can be supported within fan-in restrictions.

\begin{table}[!htb]
\centering
\begin{tabular}{@{}ccccccccccc@{}}
\toprule
Size & Reg. & Split & \multicolumn{2}{c}{Accuracy}   & \multicolumn{2}{c}{AUROC}      & \multicolumn{2}{c}{AUPRC}       & \multicolumn{2}{c}{F1}         \\ \midrule
\multirow{2}{*}{32}  & Y              & Max      & 0.93 & \multirow{2}{*}{0\%}  & 0.59 & \multirow{2}{*}{-2\%} & 0.21 & \multirow{2}{*}{-3\%} & 0.19 & \multirow{2}{*}{-1\%} \\
                     & N              & Max      & 0.93 &                       & 0.61 &                       & 0.24 &                        & 0.20 &                       \\
\multirow{2}{*}{64}  & Y              & Max      & 0.95 & \multirow{2}{*}{-1\%} & 0.55 & \multirow{2}{*}{1\%}  & 0.27 & \multirow{2}{*}{-5\%} & 0.17 & \multirow{2}{*}{1\%} \\
                     & N              & Naive        & 0.96 &                       & 0.54 &                       & 0.32 &                        & 0.16 &                       \\
\multirow{2}{*}{128} & Y              & Max      & 0.96 & \multirow{2}{*}{0\%}  & 0.57 & \multirow{2}{*}{2\%}  & 0.32 & \multirow{2}{*}{2\%}   & 0.21 & \multirow{2}{*}{4\%} \\
                     & N              & Max      & 0.96 &                       & 0.55 &                       & 0.30 &                        & 0.17 &                       \\
\multirow{2}{*}{256} & Y              & Rand       & 0.89 & \multirow{2}{*}{1\%}  & 0.50 & \multirow{2}{*}{0\%} & 0.08 & \multirow{2}{*}{-1\%} & 0.08 & \multirow{2}{*}{0\%} \\
                     & N              & Rand       & 0.88 &                       & 0.50 &                       & 0.09 &                        & 0.08 &                       \\
\multirow{2}{*}{512} & Y              & Rand      & 0.96 & \multirow{2}{*}{0\%}  & 0.50 & \multirow{2}{*}{0\%}  & 0.51 & \multirow{2}{*}{1\%}   & 0.08 & \multirow{2}{*}{0\%}  \\
                     & N              & Rand       & 0.96 &                       & 0.50 &                       & 0.50 &                        & 0.08 &                       \\ \bottomrule
\end{tabular}
\caption[Overall effect of introducing hardware regularisation.]{Overall effect of introducing hardware regularisation. Scores are presented as averages and percentage differences.}\label{tab:xylo:ablate:hw}
\end{table}
Tables \ref{tab:xylo:unreg:full} and \ref{tab:xylo:unreg:split} present detection performance results for full and split models trained without hardware-aware regularisation, respectively.
% We present detection performance results for full and split models trained without hardware-aware regularization in Tables \ref{tab:unreg:full} and \ref{tab:unreg:split}, respectively, in addition to an ablation study for hardware-aware regularization in Table \ref{tab:ablate:hw}.
We see that without regularisation, the best performance comes from the smallest model, where we suspect splitting has the least effect on sub-model dynamics; there are simply fewer changes needed to enforce hardware compatibility.
Moreover, the largest model with random splitting provides the best accuracy and AUPRC scores, suggesting that in size comes some element of redundancy. However, performance is still poor in this case.
\begin{table}[!htb]
\centering
\begin{tabular}{@{}ccccccccc@{}}
\toprule
Model Size & \multicolumn{2}{c}{Accuracy}& \multicolumn{2}{c}{AUROC}& \multicolumn{2}{c}{AUPRC}& \multicolumn{2}{c}{F1}\\ \midrule
8          & 0.9887         &0.006& 0.7589             &0.027& 0.8845            &0.078& 0.8447         &0.103\\
32         & 0.9905                &0.001& 0.9301            &0.006& 0.8829            &0.006& 0.8765         &0.007\\
64         & 0.9986               &0.000& 0.9880            &0.001& \textbf{0.9829}          &0.003& \textbf{0.9823}         &0.003\\
128        & \textbf{0.9996}              &0.000& \textbf{0.9882}           &0.002& 0.9755          &0.005& 0.9753         &0.005\\
256        & 0.9909          &0.001& 0.9128        &0.016& 0.8511            &0.019& 0.8475        &0.021\\
512        & 0.9920             &0.003& 0.9419            &0.018& 0.9001           &0.031& 0.8977        &0.032\\ \bottomrule
\end{tabular}
\caption[Detection performance results for full-sized unregularised models of varying sizes on the HERA dataset.]{Detection performance results for full-sized unregularised models of varying sizes on the HERA dataset. Scores presented as mean and standard deviation over 10 trials and best scores are bolded.}
\label{tab:xylo:unreg:full}
\end{table}

\begin{table}[!htb]
\centering
\begin{tabular}{@{}cccccccccc@{}}
\toprule
\begin{tabular}{@{}c@{}}Model\\Size\end{tabular} & \begin{tabular}{@{}c@{}}Splitting\\Method\end{tabular} & \multicolumn{2}{c}{Accuracy}& \multicolumn{2}{c}{AUROC}& \multicolumn{2}{c}{AUPRC}& \multicolumn{2}{c}{F1}\\ \midrule
\multirow{3}{*}{32}  & Maximal          & 0.9272               &0.010& \textbf{0.6109}            &0.075& 0.2444             &0.095& \textbf{0.2036}         &0.063\\
                     & Naive            & 0.9417          &0.012& 0.5334        &0.060& 0.1278      &0.089& 0.0983       &0.034\\
                     & Random           & 0.9213            &0.019& 0.5326        &0.021& 0.1174          &0.037& 0.1090     &0.023\\
\multirow{3}{*}{64}  & Maximal          & 0.9441             &0.012& 0.5213            &0.017& 0.1668           &0.083& 0.1164          &0.029\\
                     & Naive            & 0.9584           &0.002& 0.5413         &0.015& 0.3213          &0.047& 0.1551       &0.031\\
                     & Random           & 0.9530          &0.005& 0.5291     &0.012& 0.2965          &0.058& 0.1271        &0.022\\
\multirow{3}{*}{128} & Maximal          & 0.9567           &0.003& 0.5492       &0.019& 0.2962           &0.083& 0.1665         &0.050\\
                     & Naive            & 0.9587              &0.003& 0.5118            &0.007& 0.3241        &0.059& 0.0802        &0.013\\
                     & Random           & 0.9562           &0.006& 0.5279         &0.027& 0.3033        &0.093& 0.1110        &0.048\\
\multirow{3}{*}{256} & Maximal          & 0.5972               &0.126& 0.4093         &0.060& 0.1301          &0.043& 0.0782        &0.002\\
                     & Naive            & 0.8640           &0.043& 0.4938           &0.031& 0.0829           &0.035& 0.0827      &0.017\\
                     & Random           & 0.8772             &0.031& 0.5018         &0.015& 0.0866            &0.023& 0.0798       &0.006\\
\multirow{3}{*}{512} & Maximal          & 0.9259             &0.039& 0.4915        &0.010& 0.2092          &0.196& 0.0775        &0.003\\
                     & Naive            & 0.9591              &0.001& 0.5006           &0.001& 0.4692            &0.061& 0.0786      &0.003\\
                     & Random           & \textbf{0.9596}           &0.001& 0.5003           &0.000& \textbf{0.5028}  &0.033& 0.0776       &0.002\\ \bottomrule
\end{tabular}
\caption[Detection performance results for unregularised split models of varying sizes and splitting approaches on the HERA dataset.]{Detection performance results for unregularised split models of varying sizes and splitting approaches on the HERA dataset. Scores presented as mean and standard deviation over 10 trials, and best scores are bolded. We see a different trend compared to the regularised model splitting where the best scores either come from a randomly split large model, or maximal split smaller model.}
\label{tab:xylo:unreg:split}
\end{table}
\subsection{Baseline Comparison}
Finally, Table \ref{tab:xylo:hera:baselines} compares the performance of our best-performing full-width and split models, our 8-channel Xylo-width model, and prior works.
While split models fall short of SoTA performance, the surprising effectiveness of our 8-channel model is promising, and our unsplit 64-channel model achieves SoTA detection performance among SNN baselines and compelling performance against ANN baselines.
\begin{table}[htb!]
\centering
\begin{tabular}{@{}ccccc@{}}
\toprule
Work      & Model                   & AUROC & AUPRC & F1    \\ \midrule
\multicolumn{4}{l}{\textit{Algorithmic Baseline}} \\
\citet{mesarcik_learning_2022}       & AOFlagger        & 0.974 & 0.880 & 0.873 \\
\midrule
\multicolumn{4}{l}{\textit{ANN Baselines}} \\
\citet{vafaeisadr_deep_2020} & R-Net & 0.975 & 0.846 & 0.846 \\
\citet{yang_deep_2020} & RFI-Net  & 0.973 & 0.890 & 0.900 \\
\citet{mesarcik_learning_2022} & U-Net  & 0.975 & 0.896 & 0.902 \\
\citet{mesarcik_learning_2022}       & AutoEncoder       & 0.981 & 0.927 & 0.910 \\
\citet{pritchard_rfi_2024}           & AutoEncoder        & 0.983 & 0.940 & 0.939 \\
\citet{vanzyl_remove_2024} & RFDL & 0.994 & 0.965 & 0.944 \\
\citet{dutoit_comparison_2024} & ASPP & - & - & 0.985 \\
\citet{dutoit_comparison_2024} & RNet-7 & - & - & 0.989 \\
\citet{dutoit_comparison_2024} & RFI-Net & - & - & \textbf{0.993} \\
\midrule
\multicolumn{4}{l}{\textit{SNN Baselines}} \\
\citet{pritchard_rfi_2024}           & ANN2SNN         & 0.944 & 0.920 & 0.953 \\
\citet{pritchard_supervised_2024}    & BPTT                 & 0.929 & 0.785 & 0.761 \\
\citet{pritchard_spiking_2024}     & BPTT          & 0.996 & 0.914 & 0.907 \\
\citet{pritchard_advancing_2025}    & \begin{tabular}{@{}c@{}}Liquid State\\Machine\end{tabular} & 0.842 & 0.781 & 0.743 \\
\citet{pritchard_polarization-inclusive_2025}      & \begin{tabular}{@{}c@{}}Full\\Polarisation\end{tabular}  & \textbf{0.997} & 0.960 & 0.955 \\
\midrule
\multicolumn{4}{l}{\textit{This work (SNN)}} \\
This work & Full-8              & 0.750 & 0.891 & 0.857 \\
This work & Split-128           & 0.566 & 0.317 & 0.208 \\
 This work& Split-64& 0.549& 0.272&0.170\\
This work & Full-64              & 0.988 & \textbf{0.983} & 0.983 \\ \bottomrule
\end{tabular}
\caption[Detection results compared to baselines.]{Detection results compared to baselines. Best scores in bold. Our best split model (128-channels, maximally split) lags contemporary models, however our best performing 64-channel offers compelling results.}\label{tab:xylo:hera:baselines}
\end{table}
\section{Conclusion and Limitations}\label{sec:conc}
This article introduced a pipeline for training and deploying SNNs for the challenging task of RFI detection on resource-constrained neuromorphic hardware.
We developed a BPTT-trained SNN achieving high accuracy on a synthetic radio astronomy benchmark dataset, outperforming algorithmic and SNN baselines while remaining competitive with ANN baselines.
We then introduced `maximal splitting,' a greedy partitioning algorithm, to deploy our large, pre-trained SNNs onto resource-constrained neuromorphic platforms.
This pipeline approach allowed us to train models in snnTorch, before performing inference and deployment in SynSense's Rockpool library and Xylo 2 hardware, making use of the field's recent push towards interoperability through NIR \cite{pedersen_neuromorphic_2024}.
Our power usage estimates and measurements demonstrate the immense potential for SNNs and neuromorphic computing to contribute to spectral-temporal data processing in data-intensive fields such as radio astronomy.

While this article demonstrates a promising pipeline for deploying large SNNs for RFI detection onto neuromorphic hardware and achieves SoTA accuracy with the original full-width models, several limitations warrant discussion.

\begin{enumerate}
    \item \emph{Performance Degradation from Splitting}: The most significant limitation is the performance drop when we split large, high-performing SNNs into hardware-compatible sub-modules. While our `maximal splitting' method provides superior performance to baseline approaches, the split models do not retain SoTA accuracy or match the performance of smaller models trained directly for hardware. This indicates that such an approach may be inherently limited without the building of natural break-points within a network as sub-modules.
    \item \emph{Synthetic Benchmark Focus}: The primary SoTA results have been based on the synthetic HERA dataset. While this dataset is valuable and widely used within the literature, real-world astronomy data presents greater variability, more complex noise, and less certainty around ground-truth RFI detection for training. The true generalisability and robustness of our approaches need further validation on observational data.
    \item \emph{Hardware Platform Specificity}: We conducted power measurements exclusively on SynSense Xylo hardware. While providing concrete evidence of energy efficiency, this may vary across other neuromorphic architectures with their varying hardware constraints and capabilities.
    \item \emph{Feedforward Architecture Constraint}: For simplicity in model splitting and hardware deployment, this work focused on fully-connected feed-forward SNN architectures. This choice may limit learning capacity and the efficacy of model splitting without the ability to pre-define sub-modules.
    \item \emph{Temporal Dynamics for Xylo Inference}: While latency encoding was used, achieving sufficiently fine-grained temporal dynamics for optimal inference on the xylo platform after NIR conversion presented challenges. When the SNN is trained around fine-grained timings, any perturbation in the underlying neuron mode could affect overall performance.
    \item \emph{Supervised Learning Paradigm}: The core RFI detection method relies on supervised learning, which, while effective, requires labelled datasets. In radio astronomy, unsupervised, self-supervised, or even semi-supervised learning would be more desirable.
    \item \emph{Limited Power Consumption Comparisons}: While we present power usage for neuromorphic hardware, a comparison to operational RFI detection pipelines is missing. While comparison is likely to be favourable, without a concrete exploration we cannot make any substantial claims.
\end{enumerate}
Building on these limitations and the successes of our work, several promising avenues for future research emerge:
\begin{enumerate}
    \item \emph{Advanced Model Splitting and Hardware-Aware Training}: Developing more sophisticated model splitting techniques combined with retraining of sub-models, including pre-defined submodules and additional hardware-aware regularisation offline, would improve performance and efficiency post-splitting.
    \item \emph{Validation on Real-World RFI Data}: Extending the evaluation to diverse, real-world radio astronomy data (such as LOFAR, MeerKAT, ASKAP telescopes) is essential to assess the true utility and robustness of proposed methods.
    \item \emph{Exploration of Advanced SNN Architectures}: Integrating more complex SNN designs and connection types (e.g. residual, recurrent, and convolutional) into the base models and splitting algorithm approaches could enhance RFI detection and minimise accuracy drops introduced by model splitting.
    \item \emph{Optimising Temporal Encoding and Hardware Mapping}: Research into regularisation and compilation techniques to better preserve the fine-grained temporal dynamics during conversion between high-level library platforms and onto hardware could help bridge the performance gap.
    \item \emph{Semi, Self, or Un-Supervised Approaches}: Exploring other learning paradigms, such as auto-encoder-based anomaly detection schemes, could produce more robust RFI detection models better suited for real telescope deployment.
    \item \emph{Hardware Design Search}: Inverting the model splitting approach to instead search through hardware designs or configurable architectures to find more easily accommodated SNN models.
    \item \emph{Alternate Application Domains}: Exploring the applicability of the techniques this work presents to other spectro-temporal data processing domains, such as radar detection or seismic analysis, may extend the reach of SNN-based applications. It is important however to note the potential negative societal impacts of applying SNN-based techniques in dual-use tasks.
\end{enumerate}
Addressing these areas of concern will be key to advancing the practical application of SNNs and neuromorphic computing for RFI detection and more general data-intensive spectro-temporal data processing with neuromorphic computers.

Neuromorphic computing has long been envisioned as a solution for complex in-sensor processing challenges.
Radio telescopes, as some of the world's largest and most complex sensors, present a prime application domain.
Casting RFI detection as a time-series problem suitable for real-time SNN processing brings neuromorphic computing closer to the realm of sub-watt supercomputing.

\subsection*{Funding}
This work was supported by a Westpac Future Leaders Scholarship, an Australian Government Research Training Program Fees Offset and an Australian Government Research Training Program Stipend.

\subsection*{Roles}
N.J.P. was involved with conceptualisation, data curation, funding acquisition, investigation, methodology, software and writing.
A.W., R.D. and M.B. provided supervision.
R.M. provided resources.
All authors reviewed and edited the manuscript.

\subsection*{Data}
Dataset is available online: \url{https://zenodo.org/records/14676274}. Code is also available online \url{https://zenodo.org/records/17576919}.

\printbibliography

\end{document}